\documentclass[conference]{IEEEtran}

\usepackage{cite}

\ifCLASSINFOpdf
\usepackage[pdftex]{graphicx}
\usepackage{placeins}
 \DeclareGraphicsExtensions{.pdf,.jpeg,.png}
\else
\fi
\usepackage{amsmath}
\begin{document}
%
\title{Spatio-temporal Learning with Arrays of Analog Nanosynapses}
\author{\IEEEauthorblockN{
Christopher~H.~Bennett,
Damien~Querlioz, 
Jacques-Olivier~Klein 
}

\IEEEauthorblockA{C2N, Univ. Paris-Saclay, Universit\'e Paris-Sud, CNRS, 
91405 Orsay, France\\ 
Email: christopher.bennett@u-psud.fr}
}
\maketitle

\begin{abstract}
Emerging nanodevices such as resistive memories are being considered for hardware realizations of a variety of artificial neural networks (ANNs), including highly promising online variants of the learning approaches known as reservoir computing (RC) and the extreme learning machine (ELM).
We propose an RC/ELM inspired learning system built with nanosynapses that performs both on-chip projection and regression operations.
To address time-dynamic tasks, the hidden neurons of our system perform spatio-temporal integration and can be further enhanced with variable sampling or multiple activation windows. 
We detail the system and show its use in conjunction with a highly analog nanosynapse device on a standard task with intrinsic timing dynamics- the TI-46 battery of spoken digits. The system achieves nearly perfect ($99\%$) accuracy at sufficient hidden layer size, which compares favorably with software results. In addition, the model is extended to a larger dataset, the MNIST database of handwritten digits. By  translating the database into the time domain and using variable integration windows, up to $95\% $ classification accuracy is achieved. In addition to an intrinsically low-power programming style, the proposed architecture learns very quickly and can easily be converted into a spiking system with negligible loss in performance- all features that confer significant energy efficiency.
\end{abstract}


\section{Introduction}
Reservoir computing (RC) systems refer to learning models  which use the combination of random weights and recurrent projection spaces- often called the liquid state- to generate rich outputs that can be used in spatiotemporal classification or processing tasks. Previously referred to as echo state networks (ESN) as well as liquid state machines (LSM) \cite{maass2002real,jaeger2002tutorial}, these systems use linear regression or ridge regression on the output of the reservoir to analytically obtain output weights. Once these weights are set, the system can be used for on-line classification or clustering tasks. An experimental demonstration of the non-linear mapping abilities of this approach has been made in optoelectronic systems \cite{paquot2012optoelectronic}, amongst other media; however, the optimal realization of reservoir and read-out weights is seldom discussed in the RC literature. Meanwhile, the advent of non-linear nanoscale electronic elements with volatile and non-volatile modes, often called memristive devices, has paved the way towards dynamic weights with intrinsic plasticity (adaption) features \cite{yang2013memristive}. The use of said devices, hereafter ``nanosynapses'', to construct and enhance 
reservoir computers 
was first proposed discussed in \cite{kulkarni2012memristor}; here,  topology was arbitrary (randomly generated) and training employed a genetic algorithm. In  \cite{burger2015hierarchical}, the performance of an ensemble of such random reservoirs was discussed, yet an on-chip friendly read-out and regression scheme was lacking.  A completely on-chip feasible design for RC was proposed using a ring topology in \cite{kudithipudi2015design}, yet nanosynapses were only used in the output or regression layer (reservoir synapses were built with CMOS). Our proposed scheme uses memristive synapses in both layers by coupling two  crossbar arrays; this regular topology is currently being experimentally realized in both two (planar) and three (stacked) dimensions, and is straight-forward to read and write \cite{adam20173}. In contrast with RC systems, and more like a complementary approach known as the Extreme Learning Machine (ELM) \cite{huang2006extreme}, our system is feed-forward; however, even with such a simplification, strong performance on tasks with time-dynamic can be achieved by exploiting time-dynamics within the hidden layer. This approach is reminiscent of a dendritically-inspired architecture proposed by Tapson \cite{tapson2013synthesis}, who used a variety of synaptic kernels to achieve promising results on spatio-temporal tasks. Ours uses simpler spatio-temporal integration schemes specifically designed for on-chip learning with emerging, highly analog nanosynapse models. Given well-known advantages of the RC/ELM paradigm  (fast training, online operation) and the spatio-temporal data processing challenges inherent in an upcoming era of distributed computing, we expect this proposal should be of great interest to neuromorphic designers.

\section{Architecture}
\subsection{Conceptual Depiction}


A conventional RC architecture, visible in Fig.~1(a) is built with a sparsely connected graph of $m$ excitatory and inhibitory neurons in the liquid, some or all of which are recurrent; this graph is excited along a set of input channels connected to $W_\text{in}$ and feeding out along weights $W_\text{out}$. As $W_\text{in}$ is static,  $W_\text{out}$ can be solved as the pseudo-inverse of all collected output activations ($A^+$), given  a matrix of labels or expected values ($Z$), as depicted in the bottom of Fig. ~1(a). 

Our proposed scheme,
introduced in Fig. 1(b), 
also sets input weights and linearly regresses to achieve output weights, but now has a feed-forward fully-connected architecture to make it crossbar-compatible. 
To achieve the spatio-temporal features needed to emulate reservoir performance, the hidden neurons
each possess a stateful value which corresponds to the past activations they have received within the present example. 
Symbolically, given  example $s$ presented over $k$ total time frames $f$, neuron $m$ computes output as 

\begin{equation}
O_{m,s} = sign ( \sum_{f=1}^{k} \sum_{i=1}^{n} X_{i,f} W_{i,m})
\end{equation}

where $n$ is the  input dimensionality (total number of channels) and $i$ is the index of the  input channel.

The $sign$ function means that, like an RC system, neurons have binary states: excitatory (+) or inhibitory (-). The scheme is built to perform well on tasks presented frame-by-frame such as audio, video, or other tasks mapped to a time domain.

Two variations of the architecture are possible.
In the 'uniform' case, each hidden neuron integrates over every frame presented.
In the 'variable' case, each hidden neuron integrates only a subset of the time frames, specific to the hidden neuron:
the sum over $f$ is limited to a subset of $f$ values. 

This scheme is illustrated in Fig. ~1(b), where the corresponding lettered hidden layer neurons integrate over the frame areas pictured below them.

\begin{figure}[!h]
\centering
\includegraphics[width=\columnwidth]{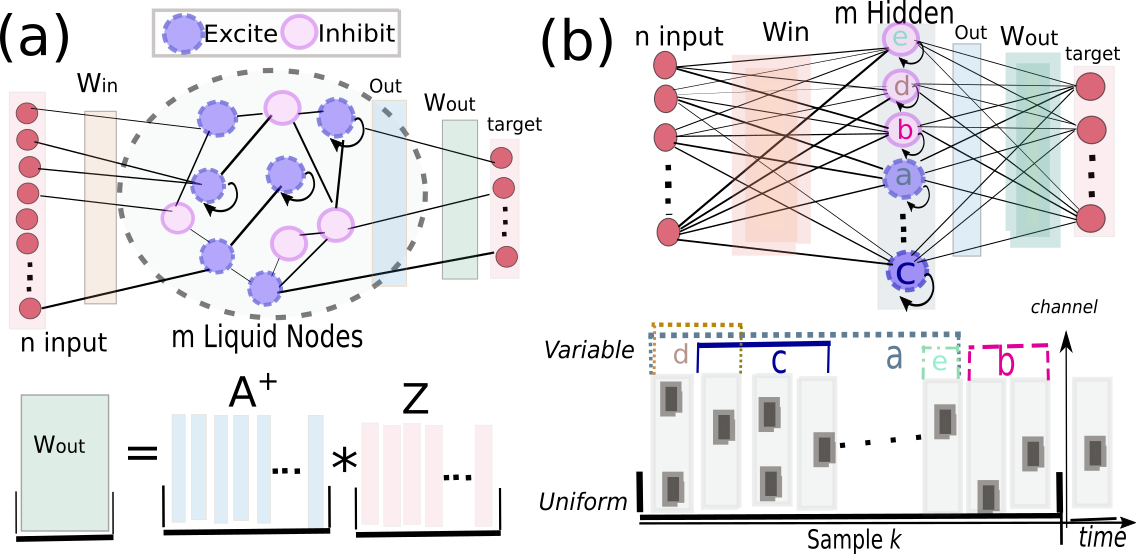}
\caption{Conceptual illustrations of (a) the standard RC/LSM paradigm and the on-chip, (b) RC-inspired scheme proposed here. }
\label{scheme}
\end{figure}

\subsection{Nanoelectric Implementation}
Our architecture implementation with nanoelectronics is presented in Fig.~2. 
It is constituted by two nanodevices crossbars implementing a ``projection layer'' and a ``readout layer'' connected by circuits implementing the hidden neurons. 

\subsubsection{Input and projection layer}
Inputs are presented to the first layer as consecutive frames of voltage pulses to $n$ input channels. 
Weights in this layer are set randomly at the initialization of operation and unchanged.
Two modes were considered: analog and spiking. Following dissemination through the crossbar, output currents at the other side are subtracted in pairs (so as to achieve negative weights) and translated through standard op-amp circuitry into the voltage domain.
Only the binary value of this voltage, or its sign are relevant to the following cell.

\subsubsection{Hidden neurons activation}


Different technological options are possible for implementing the time-integration performed by hidden neurons circuits. 
It  can be achieved digitally using counters, or in an analog fashion using capacitors, as is often done in the neuromorphic electronics field \cite{indiveri2011neuromorphic}. 
It could also potentially performed using analog nanosynapses. 

As mentioned above, each hidden  neuron integrates activation over either all frames (uniform scheme) or smaller, random subsets of frames (variable scheme). To realize the variable scheme a small memory needs to be associated and configured with each hidden neuron.  In addition to standard memory components, another crossbar of nanosynapses set randomly on or off could be used to constitute the physical reference for a sparsity matrix; in this case, dot-product operations between layers of a 3D crossbar array would significantly reduce required overhead. 


\subsubsection{Readout layer}
The readout layer uses the weight space of pairs of 
analog nanosynapses to provide on-chip regression corresponding to the activations from the hidden neurons. The scheme proposed for the readout layer is described in detail in \cite{lin2016physical}. It consists in 
a conditional pulse programming scheme which uses the combination of $+/- V_\text{pre}$ pulses   (from the hidden neurons) and $+/- V_\text{post}$ pulses (from a training cell) to implement an on-chip friendly (digital) version of stochastic gradient descent, reducing error on all device pairs in the readout layer. As pictured in Fig.~\ref{scheme},  this scheme simply checks the sign of the sample's label against the sign of the output for that class (crossbar column).  This scheme is generic to several classes of nanosynapses and highly resistant to device imperfections \cite{chabi2015chip}. The training cell can use stateful memristive devices to correctly route programming pulses to output neurons with the  corresponding  error case in the readout crossbar layer \cite{chabi2015ultrahigh}; the chosen nanosynapse has been directly integrated in such a scheme \cite{bennett2015supervised}.

\begin{figure}[!h]
\centering
\includegraphics[width=\columnwidth]{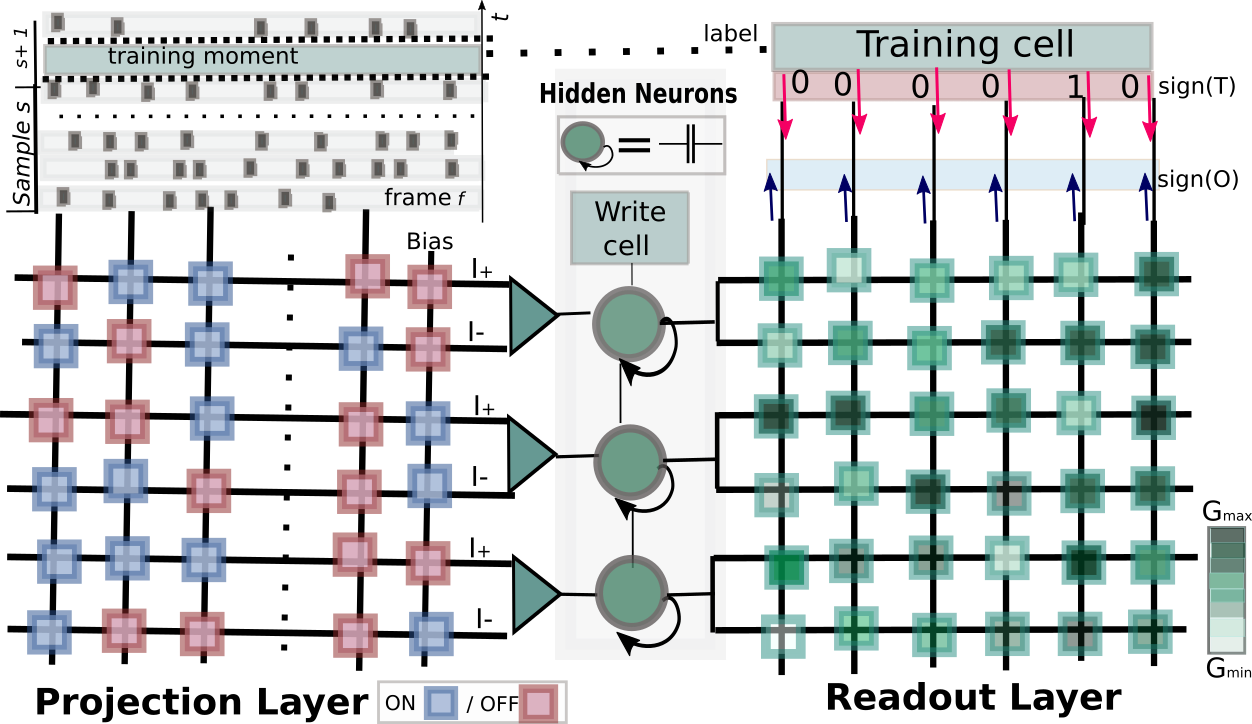}
\caption{Schematic illustrating the simulated nano-system used to obtain results. Each square icon in both cartoons represents a nano-synapse. In the first 
layer 
(projection) 
either binary device and/or natural dispersion around the maximal or minimal values of the device can be used; in the second 
layer
(readout), 
full weight range is exploited  in order to achieve on-chip regression using a simplified version of the Widrow-Hoff or delta learning rule. }
\label{scheme}
\end{figure}

\subsection{Nanosynapse}
The nanosynapse model is extrapolated from experimental data obtained from the hardware learning of a highly analog polyermic nanosynapse with several hundred writable/readable states. The device physically functions via active filamentary redox mechanisms occurring within an electrografted polymer thin-film of tris-bipyridine iron complexes (TBFe) \cite{lin2016physical}. Our generic model evaluates the evolution of the device between $G_\text{max}$ and $G_\text{min}$; conductance increases if and only if it is biased at a voltage $V_b$ above a critical threshold $V_{th1}$ but below $V_{th2}$ and decreases above $V_{th2}$, as a function of its total addressable states $g$ and the write range , $r = G_\text{max} - G_\text{min}$, as follows : 
 \begin{equation}
  \frac{dG}{dt} = \begin{cases}
         V_{th2} > |V_{b}|>V_{th1} &\text{$f(|V_{b1} - V_{th1}|,g,r)$} \\
        |V_{b}| > V_{th2} &\text{$-f(|V_{b2} - V_{th2}|,g,r)$} \\
        |V_{b}| < V_{th1} &\text{0}.
        \end{cases}
  \label{memeq}
 \end{equation}
This case is significantly simplified by using a constant programming scheme where $V_{b2} = V_p+ = 5.5V$ for reset, $V_{b1} = V_p- = 3.3V$ and programming (pulse) length $dt = 100 \mu s$, such that $ \Delta G = f(g,r)$ can be approximated in a discrete time-step simulator during each writing (learning) phase. Two varieties of this model were considered on the network level: perfect devices, for which every device has identical extrema ( $G_\text{max} = 69.5 \mu S, G_\text{min} = 2.1 \mu S$) and identical thresholds; and imperfect devices, which have extrema distributed around a Gaussian spread from those means, and with first and second thresholds distributed around $V_\text{th1} = 3.1$, $V_\text{th2} = 5.5$. For all the following simulations, 7 bits or $g=128$ addressable states were considered for the 
analog readout layer. 
For the projection layer ($W_\text{in}$) which is unadaptive, the devices were either in 'binary mode' and 
programmed 
randomly at one extrema (perfect devices); or were all 
programmed  
randomly around the same extrema (ON or OFF), taking advantage of the natural Gaussian curve as proposed in \cite{ELMMemristor} (imperfect devices).


\section{Simulation Methodology}


First, we considered a sub-set of 500 examples from the TI-46 corpus of spoken digits. These spoken digits were encoded using an artificial cochleagram; pre-processing was based on the passive ear model or filter bank designed by R.F. Lyon \cite{lyon1982computational}. Full details on this pre-processing and database are available in \cite{verstraeten2005isolated}. The processed data we used in our tests shows 77 channels with between 50 and 100 time frames or steps during each example digit.  The dataset was imported from and used in the context of a generic software simulator for reservoir computing written in Python (Oger) \cite{verstraeten2012oger}. The identical database and pre-processing methods were  used to present this set as training and testing examples to a custom discrete-time crossbar simulator which tracks the evolution of all hidden neurons activation values and (second layer) device weights; this software was also written in Python.

The 500 samples were divided into a training set of 350 samples, and a test set of 150 samples. The training set is labeled and used either to teach the readout layer or to build an output matrix for ridge regression; conversely, testing examples are presented label-less, and the prediction or inference is compared externally to the system. Each full training procedure  consisted of 10 subsequent presentations (epochs) of the sample set (5k iterations); examples were always shuffled between each epoch.


Although this task has intrinsic time dynamic, the number of total samples is very small. To further estimate the generalization abilities of our architectures, we also tested with the well known MNIST database of handwritten digits \cite{lecun2010mnist}, which consists of 70,000 samples (10,000 tests, 60,000 training examples).
In order to lend time-dynamic to the task we presented all individual examples (from training and test) as 28 frames presented subsequently to the input layer, as in \cite{burger2014volatile}. As in the TI-46 database, training examples were always shuffled (presented in a different order). Training with this database, due to its large size, consists of only a single epoch; no distortions or pre-processing were made before presentation; reported scores are always based on classification on the entire test database.

For analog mode presentation:  pixel intensity or channel's analog value for cochleagram were input directly as the corresponding voltage pulses. For spiking mode, all negative cochleagram values (range $[-1,1]$) were converted to 0 (no spike), and all positive to 1 (spike); for MNIST, pixel luminosity greater than 0.5 spiked, and that below did not (range $[0,1]$). When applying noise to spike channels, the noted percent of pixels or channels at each moment flipped from one bit value to the other (on average).


\section{Learning Performances}
\subsection{Effect of Liquid/ Hidden Layer size}
\begin{figure*}[!t]
\centering
\includegraphics[width=7.2in]{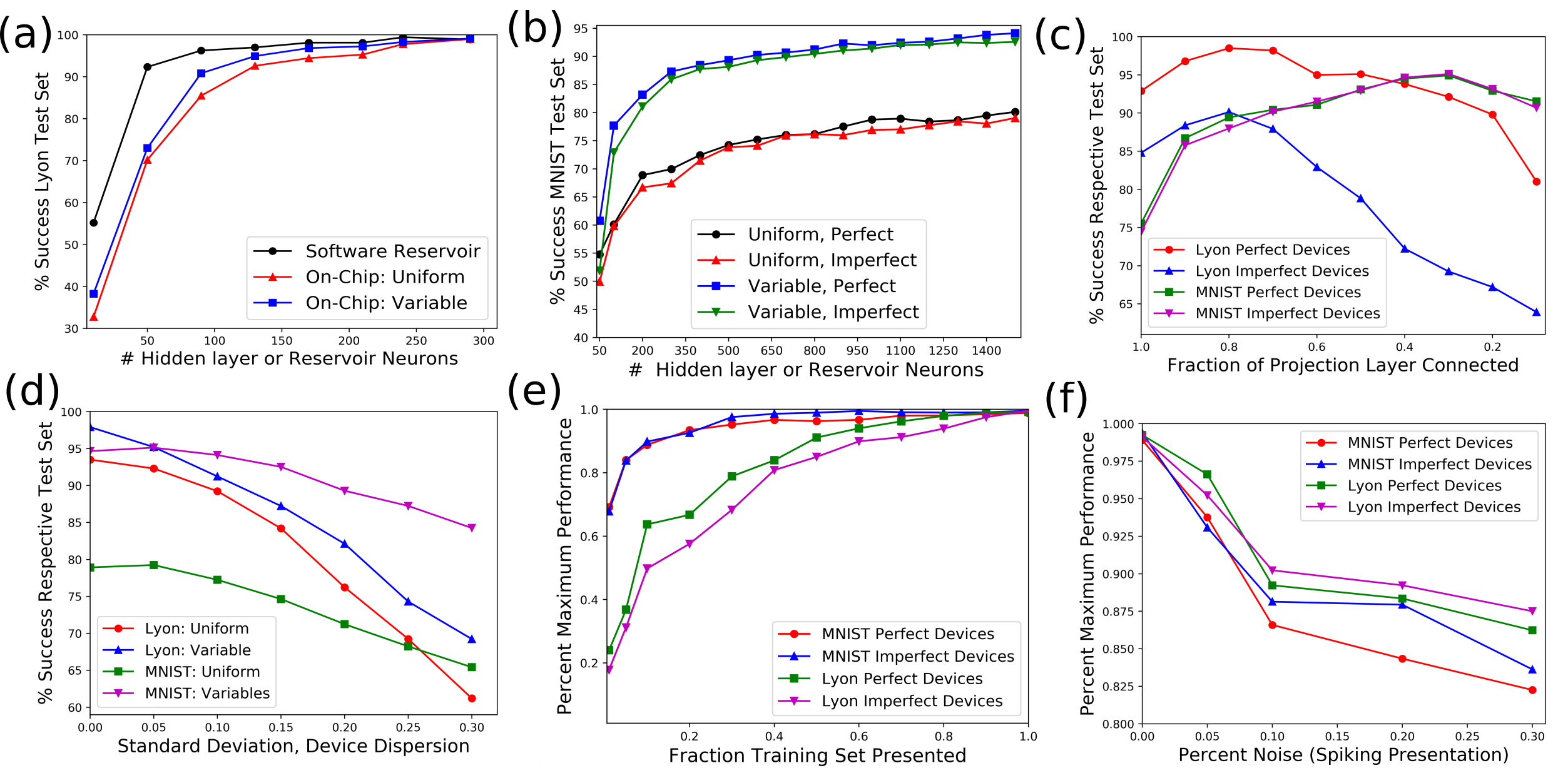}
\caption{ Performance of the considered architecture on the two chosen tasks. (a) shows performance on Lyon for software, on-chip simple integration, and on-chip sparse integration; (c) shows both considered integration schemes for the MNIST challenge where mild variability ($\sigma =10\%$)  is showed in  the 'imperfect' series; (c) shows deterioration in performance as dispersion the $G_\text{on}$, $G\text{off}$ parameters is increased; (d) shows response of variably integrating systems to the degree of spareness in the first layer, where 1.0 is the default (uniform) case; (e) shows convergence speed  of the considered systems as a fraction of the total given samples in other graphs; (f) shows degradation in performance as the input channels to the first layer are increasingly simplified (analog to digital) and corrupted (noise). }
\label{results}
\end{figure*}

As visible in Fig.~\ref{results} (a), both the software reservoir and our proposed crossbar interpretation are strongly dependent on a large enough number of nodes in the reservoir or hidden layer to reach robust performance. However, while the software reservoir achieves $ > 95\%$ already by $m=100$ neurons, a dimensionality of  at least $m=200$ is required for the crossbar based version when it is built with perfect nanosynapses. As visible, the variable sampling method creates a noticeable but not dramatic improvement.

The case of the MNIST challenge shows the same strong dependence of performance on hidden layer size, but in this case, the performance improvement between the uniform integration and variable schemes is dramatic. The uniform case only approaches  $80\% $ at large number of hidden neurons; this is an inferior result for a two layer system, given that a single layer (perceptron) system can achieve $88\% $ in software \cite{lecun1998gradient} and a one-layer crossbar system using the same device can approximate this result \cite{lin2016physical}. This suggests that the simple, or uniform integration scheme that was used earlier is not adequate to differentiate between many presented MNIST samples in the training phase. Two factors may be at play: first, when artificially converted into the time domain few input channels are available in general; second, integration (summation) across the entire time slice cannot enhance dimensionality but only reduce it. This would be especially punishing in the MNIST case due to the complexity of the training data (which is not, in general, linearly separable).
In contrast, adding hidden layer variability results in a substantial boost in performance, with ultimate performance at 95\% at large hidden size. To obtain these results a random subset of one fourth of the frames from each image was selected by each neuron; this preference is consistent over all presentations and between training and testing.



\subsection{Effect of sparsity on variable scheme}
Next, we explored whether optimal sparse connection levels exist. As visible in Fig.~\ref{results} (c), the optimal sparsity greatly depends upon the task; the optimal sparsity for MNIST is quite high, with best performances coming when between 40-20\%  of frames are active per neuron. For the Lyon task, 85-65\% connection produces the best results. While outcome for MNIST devices with 10\% variability is mostly equivalent to perfect devices, the same cannot be said for Lyon imperfect devices which trail performance relative to perfect ones increasingly as sparsity increases.

\subsection{Effect of readout device variability}
While projection layer variability can easily be taken in stride by using normal distributions as an asset, imperfect devices in the adaptive (readout) layer can confound the learning rule's performance. Fig.~\ref{results} (d) shows performance on each of the two considered challenges as a function of increasing standard deviation or dispersion of maximal, minimal conductances and thresholds at $M=200$ for the  Lyon task, and $M=1200$ for the MNIST task. As visible, slight variability is not very harmful, while greater than $10\%-15\%$ starts to substantively effect performance. Notably, the rate of deterioration on the Lyon task is much higher than the MNIST task; in the far smaller readout crossbar for the Lyon task, each device is  more important to successfully resolving the problem. Reducing the impact of device imperfections in general could be realized via better engineering or through architectural tweaks such as using more than two nanodevices per single nanosynapse element.

\subsection{Fast Training Capabilities}
One of the signature benefits of the RC/ELM paradigm is that such systems generally require only a fraction of the quantity of training examples needed to solve the problem in more computationally demanding methods; \textit{e.g.} multi-layer backpropagation or convolutional neural networks  often require dozens of epochs (millions of individual updates) to properly converge. As illustrated in Fig. \ref{results} (e), the Lyon and MNIST tasks achieve within 10\% of the maximum performance in as few as 20\% (12k) training examples, in the former case, and 60\% (2.1k) training examples, in the latter case. For even more approximate applications (less stringent accuracy requirements), 80\% of performance can be reached in only 5k samples, for the case of the MNIST, and 1.4k samples, for the Lyon case. As small error rates lead to substantial energy savings, these results suggest the system is an attractive approximate computer.

\subsection{Performance with Spiking Input}

In neuromorphic applications, spike coding bestows extreme energy efficiency \cite{merolla2014million}. We tested the capability of our system to still provide strong solutions when the problem was presented in either in  perfectly presented spike form, or imperfect (increasingly noisy) spiking channel input, as described previously. As visible in Fig.~\ref{results} (f), the initial conversion from analog to digital (spiking) presentation has a slim 
effect on ultimate performance (between 1.1\% and 0.4\% drop from ideal analog case). With increasing noise, performance deteriorates slowly; on a slightly noisy channel ( 5\% bit flips) performance drops only 4-8\%, and even with nearly one third of the channel corrupted  performance is still within 80\% of the maximum. The system's performance degradation on the Lyon task is less than that on the  MNIST challenge; this shows that encoding on the latter on only 28 channels is intrinsically difficult and with added noise many of the features are lost.

\subsection{Energy Estimates}
The energy benefits of programming devices in a crossbar array are substantial relative to the alternatives such as SRAM \cite{agarwal2015energy}. Even better, this scheme keeps the first layer weights static and trains the second layer using a conditional programming pulse scheme (not every output neuron is programmed at every training step). Due to high voltage levels for the simulated TBFe polymeric device ($V_\text{p}=4.4V$), we also include energy estimates using an alternate polymeric electrochemical nanosynapse called ENODe \cite{van2017non} where  $V_\text{p}=0.5mV$. However, there is a trade-off in programming speed, since $t_\text{prog}=100 \mu s$ for TBFe and  $t_\text{prog}=2 s$ for ENODe. The  energy cost per elementary write operation for the two devices is computed finally as $0.077 \mu J$ and  $0.325 nJ$, respectively. Total programming expenditure for the Lyon task and the MNIST task for both polymeric nanosynapses at full training (full performance), within 10 \% of max performance, and within 20\% of max performance  following the analysis in Fig. \ref{results} (e) are visible in Table \ref{tab:energy}; system (hidden layer) size for Lyon is $M=200$, and $M=1200$ for MNIST. While  this estimation is only raw programming cost, power draw for sum/ integration operations in the hidden layer, and checking/routing of error should be far less, especially if they use nanosynapses.
\begin{table}[t]
\centering
\begin{tabular}{|l|c|c|c|c|c|}
\hline
 Device/Task & Full Performance & 10\% Loss & 20\% Loss \\
\hline
ENODe-Lyon & 0.65mJ & 0.39mJ & 0.182mJ \\
TBFe-Lyon & 154mJ & 92.4mJ & 43.12mJ \\
TBFe-MNIST & 11.09J & 2.22J & 0.924J \\
ENODe-MNIST & 46.8 mJ& 9.36mJ & 3.9mJ \\
\hline
\end{tabular}
\caption{Programming cost for two possible nanosynapses}
\label{tab:energy}
\end{table}

\subsection{Benchmarks}
Our  best obtained results on the Lyon task match not only RC software results, but the best obtained results with  complex Long-short term memory cell (LSTM) networks \cite{graves2004biologically}. Our results are very slightly inferior to physically realized reservoir computers; \cite{paquot2012optoelectronic} obtained only 0.4\% error.  Finally, to test how critical the two layer design is to the high performance we achieved, we simulated a one-layer system with the training data directly presented (only the readout layer). Using this approach, we achieved only 56\% using the uniform and 62\% using the variable spatio-temporal integration scheme. This suggests that direct integration is insufficient for effective spatio-temporal learning. As suggested in \cite{tapson2013synthesis}, the critical operation expressed by synaptic (in our system, hidden layer)  kernels is to project inputs into a higher dimensional space. 

Many benchmarks are available for MNIST, but of more relevance to our case are those using nanosynapse weights in part or whole.  In comparison to the results obtained in \cite{burger2014volatile}, which again presented that database similarly to memristive crossbars, we obtained superior results while also avoiding off-chip computational requirements. Our top obtained result is superior to the standard ELM result (91.5\%) and similar to the 'enhanced ELM' result (94.1\%) showed in \cite{bennett2016exploiting}. But unlike the latter case, our system requires no special priming/pre-training step in the first layer. A similar combination of unsupervised and supervised learning was used in  \cite{querlioz2012bioinspired} to achieved $93.5\%$ accuracy on the MNIST test set; in contrast, our system avoids relying on any first layer plasticity effects. For many, although not all, nanosynapses, plasticity learning rules such as STDP rely on carefully designed waveforms.
 
\section{Discussion}

While the proposed  design has strong similarities to the RC paradigm: the first layer weights (synaptic attractions) are fixed, standard regression can be used to extract the value of this processing network, and neurons  show both positive (excitatory) and negative (inhibitory) activations, like ELM, the system is feedforward, not recurrent. As recurrence bestows a higher memory capacity on a reservoir, it is not surprising that  more hidden  neurons are required in our case than a true reservoir. The  computational capabilities of RCs and ELMs have previously been compared in \cite{butcher2013reservoir}, which noted that while RCs imply an unfortunate trade-off between non-linearity (mapping performance) and memory capacity (ability to retain said map beyond the fading time window), a synthesis of ELM and RC approaches might overcome this dilemma. Our system implicitly addresses this trade-off by forcing non-linearity at a large set of hidden neurons, and preserving capacity through stateful variables. 

Our results with the frame-by-frame MNIST task suggest that achieving strong performance on a high-dimensional task requires more than simple spatio-temporal integration. When hidden  neurons time variability is included so as to add diversity to activation functions, the computational power of the system amplifies substantially as neurons  specialize on their preferred frames from the database. Effectively, this approach forces greater sparsity (breaks all-to-all connectivity). Software RC systems intrinsically exploit sparsity,  while ELM systems exploiting this effect are rarer; different local receptive fields were suggested in \cite{huang2015local} and used to achieve good performance on the NORB database, but the scheme is  too complex to be easily implemented on-chip. In contrast, our proposed scheme is hardware-friendly, and has been designed with future 3D crossbar topologies in mind.


\section{Conclusion}
By integrating two standard nanoelectric structures (crossbars) with  hidden neurons  performing spatio-temporal activation/integration, we have emulated many of the features of  RC/ELM software systems: high performance, online learning, and resilience to imperfect dynamics or inputs. 99\%  on the T1-46 battery  matches the performance of a software reservoir computer with floating point weights; while the same equivalence cannot be stated about our results on the MNIST task, 95\%  is nonetheless a strong result for a crossbar based learning system. Immediate next steps in translating the proposed design into a prototype include study on the energy draw and operation of optimal CMOS associated circuits, especially those involving hidden layer functions,  electrical simulation of the proposed system to explore non-ideal effects, and design of toy problems capable of being implemented on far smaller crossbars than those used in the present study.


%

\section*{Acknowledgment}
This work was supported by the Nanodesign Paris-Saclay lidex.

\ifCLASSOPTIONcaptionsoff
  \newpage
\fi



%
\bibliographystyle{IEEEtran}
\bibliography{references}

\begin{thebibliography}{10}
\providecommand{\url}[1]{#1}
\csname url@samestyle\endcsname
\providecommand{\newblock}{\relax}
\providecommand{\bibinfo}[2]{#2}
\providecommand{\BIBentrySTDinterwordspacing}{\spaceskip=0pt\relax}
\providecommand{\BIBentryALTinterwordstretchfactor}{4}
\providecommand{\BIBentryALTinterwordspacing}{\spaceskip=\fontdimen2\font plus
\BIBentryALTinterwordstretchfactor\fontdimen3\font minus
  \fontdimen4\font\relax}
\providecommand{\BIBforeignlanguage}[2]{{%
\expandafter\ifx\csname l@#1\endcsname\relax
\typeout{** WARNING: IEEEtran.bst: No hyphenation pattern has been}%
\typeout{** loaded for the language `#1'. Using the pattern for}%
\typeout{** the default language instead.}%
\else
\language=\csname l@#1\endcsname
\fi
#2}}
\providecommand{\BIBdecl}{\relax}
\BIBdecl

\bibitem{maass2002real}
W.~Maass, T.~Natschl{\"a}ger, and H.~Markram, ``Real-time computing without
  stable states: A new framework for neural computation based on
  perturbations,'' \emph{Neural computation}, vol.~14, no.~11, pp. 2531--2560,
  2002.

\bibitem{jaeger2002tutorial}
H.~Jaeger, \emph{Tutorial on training recurrent neural networks, covering BPPT,
  RTRL, EKF and the" echo state network" approach}.\hskip 1em plus 0.5em minus
  0.4em\relax GMD-Forschungszentrum Informationstechnik, 2002, vol.~5.

\bibitem{paquot2012optoelectronic}
Y.~Paquot, F.~Duport, A.~Smerieri, J.~Dambre, B.~Schrauwen, M.~Haelterman, and
  S.~Massar, ``Optoelectronic reservoir computing,'' \emph{Scientific reports},
  vol.~2, p. 287, 2012.

\bibitem{yang2013memristive}
J.~J. Yang, D.~B. Strukov, and D.~R. Stewart, ``Memristive devices for
  computing,'' \emph{Nature nanotechnology}, vol.~8, no.~1, pp. 13--24, 2013.

\bibitem{kulkarni2012memristor}
M.~S. Kulkarni and C.~Teuscher, ``Memristor-based reservoir computing,'' in
  \emph{Nanoscale Architectures (NANOARCH), 2012 IEEE/ACM International
  Symposium on}.\hskip 1em plus 0.5em minus 0.4em\relax IEEE, 2012, pp.
  226--232.

\bibitem{burger2015hierarchical}
J.~B{\"u}rger, A.~Goudarzi, D.~Stefanovic, and C.~Teuscher, ``Hierarchical
  composition of memristive networks for real-time computing,'' in
  \emph{Nanoscale Architectures (NANOARCH), 2015 IEEE/ACM International
  Symposium on}.\hskip 1em plus 0.5em minus 0.4em\relax IEEE, 2015, pp. 33--38.

\bibitem{kudithipudi2015design}
D.~Kudithipudi, Q.~Saleh, C.~Merkel, J.~Thesing, and B.~Wysocki, ``Design and
  analysis of a neuromemristive reservoir computing architecture for biosignal
  processing,'' \emph{Frontiers in neuroscience}, vol.~9, 2015.

\bibitem{adam20173}
G.~C. Adam, B.~D. Hoskins, M.~Prezioso, F.~Merrikh-Bayat, B.~Chakrabarti, and
  D.~B. Strukov, ``3-d memristor crossbars for analog and neuromorphic
  computing applications,'' \emph{IEEE Transactions on Electron Devices},
  vol.~64, no.~1, pp. pp--312, 2017.

\bibitem{huang2006extreme}
G.-B. Huang, Q.-Y. Zhu, and C.-K. Siew, ``Extreme learning machine: theory and
  applications,'' \emph{Neurocomputing}, vol.~70, no.~1, pp. 489--501, 2006.

\bibitem{tapson2013synthesis}
J.~Tapson, G.~Cohen, S.~Afshar, K.~Stiefel, Y.~Buskila, R.~Wang, T.~J.
  Hamilton, and A.~van Schaik, ``Synthesis of neural networks for
  spatio-temporal spike pattern recognition and processing,'' \emph{arXiv
  preprint arXiv:1304.7118}, 2013.

\bibitem{indiveri2011neuromorphic}
G.~Indiveri, B.~Linares-Barranco, T.~J. Hamilton, A.~Van~Schaik,
  R.~Etienne-Cummings, T.~Delbruck, S.-C. Liu, P.~Dudek, P.~H{\"a}fliger,
  S.~Renaud \emph{et~al.}, ``Neuromorphic silicon neuron circuits,''
  \emph{Frontiers in neuroscience}, vol.~5, p.~73, 2011.

\bibitem{lin2016physical}
Y.-P. Lin, C.~H. Bennett, T.~Cabaret, D.~Vodenicarevic, D.~Chabi, D.~Querlioz,
  B.~Jousselme, V.~Derycke, and J.-O. Klein, ``Physical realization of a
  supervised learning system built with organic memristive synapses,''
  \emph{Scientific Reports}, vol.~6, 2016.

\bibitem{chabi2015chip}
D.~Chabi, W.~Zhao, D.~Querlioz, and J.-O. Klein, ``On-chip universal supervised
  learning methods for neuro-inspired block of memristive nanodevices,''
  \emph{ACM Journal on Emerging Technologies in Computing Systems (JETC)},
  vol.~11, no.~4, p.~34, 2015.

\bibitem{chabi2015ultrahigh}
D.~Chabi, Z.~Wang, C.~Bennett, J.-O. Klein, and W.~Zhao, ``Ultrahigh density
  memristor neural crossbar for on-chip supervised learning,'' \emph{IEEE
  Transactions on Nanotechnology}, vol.~14, no.~6, pp. 954--962, 2015.

\bibitem{bennett2015supervised}
C.~Bennett, D.~Chabi, T.~Cabaret, B.~Jousselme, V.~Derycke, D.~Querlioz, and
  J.-O. Klein, ``Supervised learning with organic memristor devices and
  prospects for neural crossbar arrays,'' in \emph{Nanoscale Architectures
  (NANOARCH), 2015 IEEE/ACM International Symposium on}, no. Proceedings of
  IEEE.\hskip 1em plus 0.5em minus 0.4em\relax IEEE, 2015, pp. 181--186.

\bibitem{ELMMemristor}
M.~Suri and V.~Parmar, ``Exploiting intrinsic variability of filamentary
  resistive memory for extreme learning machine architectures,'' \emph{IEEE
  Trans. Nanotechnol.}, vol.~14, no.~6, pp. 963--968, 2015.

\bibitem{lyon1982computational}
R.~Lyon, ``A computational model of filtering, detection, and compression in
  the cochlea,'' in \emph{Acoustics, Speech, and Signal Processing, IEEE
  International Conference on ICASSP'82.}, vol.~7.\hskip 1em plus 0.5em minus
  0.4em\relax IEEE, 1982, pp. 1282--1285.

\bibitem{verstraeten2005isolated}
D.~Verstraeten, B.~Schrauwen, D.~Stroobandt, and J.~Van~Campenhout, ``Isolated
  word recognition with the liquid state machine: a case study,''
  \emph{Information Processing Letters}, vol.~95, no.~6, pp. 521--528, 2005.

\bibitem{verstraeten2012oger}
D.~Verstraeten, B.~Schrauwen, S.~Dieleman, P.~Brakel, P.~Buteneers, and
  D.~Pecevski, ``Oger: modular learning architectures for large-scale
  sequential processing,'' \emph{Journal of Machine Learning Research},
  vol.~13, no. Oct, pp. 2995--2998, 2012.

\bibitem{lecun2010mnist}
Y.~LeCun, C.~Cortes, and C.~J. Burges, ``Mnist handwritten digit database,''
  \emph{AT\&T Labs [Online]. Available: http://yann. lecun. com/exdb/mnist},
  vol.~2, 2010.

\bibitem{burger2014volatile}
J.~B{\"u}rger and C.~Teuscher, ``Volatile memristive devices as short-term
  memory in a neuromorphic learning architecture,'' in \emph{Proceedings of the
  2014 IEEE/ACM International Symposium on Nanoscale Architectures}.\hskip 1em
  plus 0.5em minus 0.4em\relax ACM, 2014, pp. 104--109.

\bibitem{lecun1998gradient}
Y.~LeCun, L.~Bottou, Y.~Bengio, and P.~Haffner, ``Gradient-based learning
  applied to document recognition,'' \emph{Proceedings of the IEEE}, vol.~86,
  no.~11, pp. 2278--2324, 1998.

\bibitem{merolla2014million}
P.~A. Merolla, J.~V. Arthur, R.~Alvarez-Icaza, A.~S. Cassidy, J.~Sawada,
  F.~Akopyan, B.~L. Jackson, N.~Imam, C.~Guo, Y.~Nakamura \emph{et~al.}, ``A
  million spiking-neuron integrated circuit with a scalable communication
  network and interface,'' \emph{Science}, vol. 345, no. 6197, pp. 668--673,
  2014.

\bibitem{agarwal2015energy}
S.~Agarwal, T.-T. Quach, O.~Parekh, A.~H. Hsia, E.~P. DeBenedictis, C.~D.
  James, M.~J. Marinella, and J.~B. Aimone, ``Energy scaling advantages of
  resistive memory crossbar based computation and its application to sparse
  coding,'' \emph{Frontiers in neuroscience}, vol.~9, 2015.

\bibitem{van2017non}
Y.~van~de Burgt, E.~Lubberman, E.~J. Fuller, S.~T. Keene, G.~C. Faria,
  S.~Agarwal, M.~J. Marinella, A.~A. Talin, and A.~Salleo, ``A non-volatile
  organic electrochemical device as a low-voltage artificial synapse for
  neuromorphic computing,'' \emph{Nature Materials}, vol.~16, no.~4, pp.
  414--418, 2017.

\bibitem{graves2004biologically}
A.~Graves, D.~Eck, N.~Beringer, and J.~Schmidhuber, ``Biologically plausible
  speech recognition with lstm neural nets,'' in \emph{International Workshop
  on Biologically Inspired Approaches to Advanced Information
  Technology}.\hskip 1em plus 0.5em minus 0.4em\relax Springer, 2004, pp.
  127--136.

\bibitem{bennett2016exploiting}
C.~H. Bennett, S.~La~Barbera, A.~F. Vincent, J.-O. Klein, F.~Alibart, and
  D.~Querlioz, ``Exploiting the short-term to long-term plasticity transition
  in memristive nanodevice learning architectures,'' in \emph{Neural Networks
  (IJCNN), 2016 International Joint Conference on}.\hskip 1em plus 0.5em minus
  0.4em\relax IEEE, 2016, pp. 947--954.

\bibitem{querlioz2012bioinspired}
D.~Querlioz, W.~Zhao, P.~Dollfus, J.-O. Klein, O.~Bichler, and C.~Gamrat,
  ``Bioinspired networks with nanoscale memristive devices that combine the
  unsupervised and supervised learning approaches,'' in \emph{Proceedings of
  the 2012 IEEE/ACM International Symposium on Nanoscale Architectures}.\hskip
  1em plus 0.5em minus 0.4em\relax ACM, 2012, pp. 203--210.

\bibitem{butcher2013reservoir}
J.~Butcher, D.~Verstraeten, B.~Schrauwen, C.~Day, and P.~Haycock, ``Reservoir
  computing and extreme learning machines for non-linear time-series data
  analysis,'' \emph{Neural networks}, vol.~38, pp. 76--89, 2013.

\bibitem{huang2015local}
G.-B. Huang, Z.~Bai, L.~L.~C. Kasun, and C.~M. Vong, ``Local receptive fields
  based extreme learning machine,'' \emph{IEEE Computational Intelligence
  Magazine}, vol.~10, no.~2, pp. 18--29, 2015.

\end{thebibliography}

\end{document}